%% file: main.tex
\documentclass[letterpaper]{article} 
\usepackage{aaai24}  
\usepackage{times}  
\usepackage{helvet}  
\usepackage{courier}  
\usepackage[hyphens]{url}  
\usepackage{graphicx} 
\usepackage{natbib}  
\usepackage[font=footnotesize,labelfont=bf]{caption} 
\usepackage{algorithm}
\usepackage{algorithmic}

\usepackage{newfloat}
\usepackage{listings}
\DeclareCaptionStyle{ruled}{labelfont=normalfont,labelsep=colon,strut=off} 
\floatstyle{ruled}
\newfloat{listing}{tb}{lst}{}
\floatname{listing}{Listing}
\usepackage{tabularx, booktabs, multirow}

\title{Balancing Continual Learning and Fine-tuning for Human Activity Recognition}
\author{
Chi Ian Tang\textsuperscript{\rm 1, 2}\footnote{Corresponding author (ian.tang@nokia-bell-labs.com)}, 
Lorena Qendro\textsuperscript{\rm 1}, Dimitris Spathis\textsuperscript{\rm 1}, Fahim Kawsar\textsuperscript{\rm 1},\\Akhil Mathur\textsuperscript{\rm 1}, Cecilia Mascolo\textsuperscript{\rm 2}\\ 
}
\affiliations{
    \textsuperscript{\rm 1}Nokia Bell Labs, UK  \qquad
    \textsuperscript{\rm 2}University of Cambridge, UK

}

\usepackage{bibentry}

\begin{document}

\maketitle

\input{sections/0-abstract}
\input{sections/1-introduction}
\input{sections/3-method}
\input{sections/4-evaluation}

\input{sections/5-conclusion}

\bibliography{aaai24}
\clearpage \newpage
\appendix

\input{sections/6-appendix}
\end{document}

%% file: sections/0-abstract.tex
\begin{abstract}
Wearable-based Human Activity Recognition (HAR) is a key task in human-centric machine learning due to its fundamental understanding of human behaviours. Due to the dynamic nature of human behaviours, continual learning promises HAR systems that are tailored to users' needs. However, because of the difficulty in collecting labelled data with wearable sensors, existing approaches that focus on supervised continual learning have limited applicability, while unsupervised continual learning methods only handle representation learning while delaying classifier training to a later stage. This work explores the adoption and adaptation of CaSSLe, a continual self-supervised learning model, and Kaizen, a semi-supervised continual learning model that balances representation learning and down-stream classification, for the task of wearable-based HAR. These schemes re-purpose contrastive learning for knowledge retention and, Kaizen combines that with self-training in a unified scheme that can leverage unlabelled and labelled data for continual learning. In addition to comparing state-of-the-art self-supervised continual learning schemes, we further investigated the importance of different loss terms and explored the trade-off between knowledge retention and learning from new tasks. In particular, our extensive evaluation demonstrated that the use of a weighting factor that reflects the ratio between learned and new classes achieves the best overall trade-off in continual learning.
\end{abstract}

%% file: sections/1-introduction.tex
\section{Introduction}

The widespread adoption of mobile devices has created opportunities in human-centric computing by capturing user behaviours through sensors on devices people carry. A key challenge in user modelling is the presence of shifts in human behaviours, where user behaviours can change over time. The problem of catastrophic forgetting \cite{kirkpatrick2017overcoming, aljundi2018memory, diethe2019continual, van2019three} has been a critical challenge in continual learning, in which deep learning models forget what has been learned when being trained on data with shifted distributions. Even though many approaches have been proposed to mitigate catastrophic forgetting \cite{kirkpatrick2017overcoming, li2017learning, shin2017continual, wu2019large}, most assume abundant labelled data for every new distribution, which is unrealistic for mobile sensing. Collecting quality ground truth when data is generated on the fly in wearable-based user modelling is particularly challenging.

\begin{figure}[t]
\begin{center}
   \includegraphics[width=0.7\linewidth]{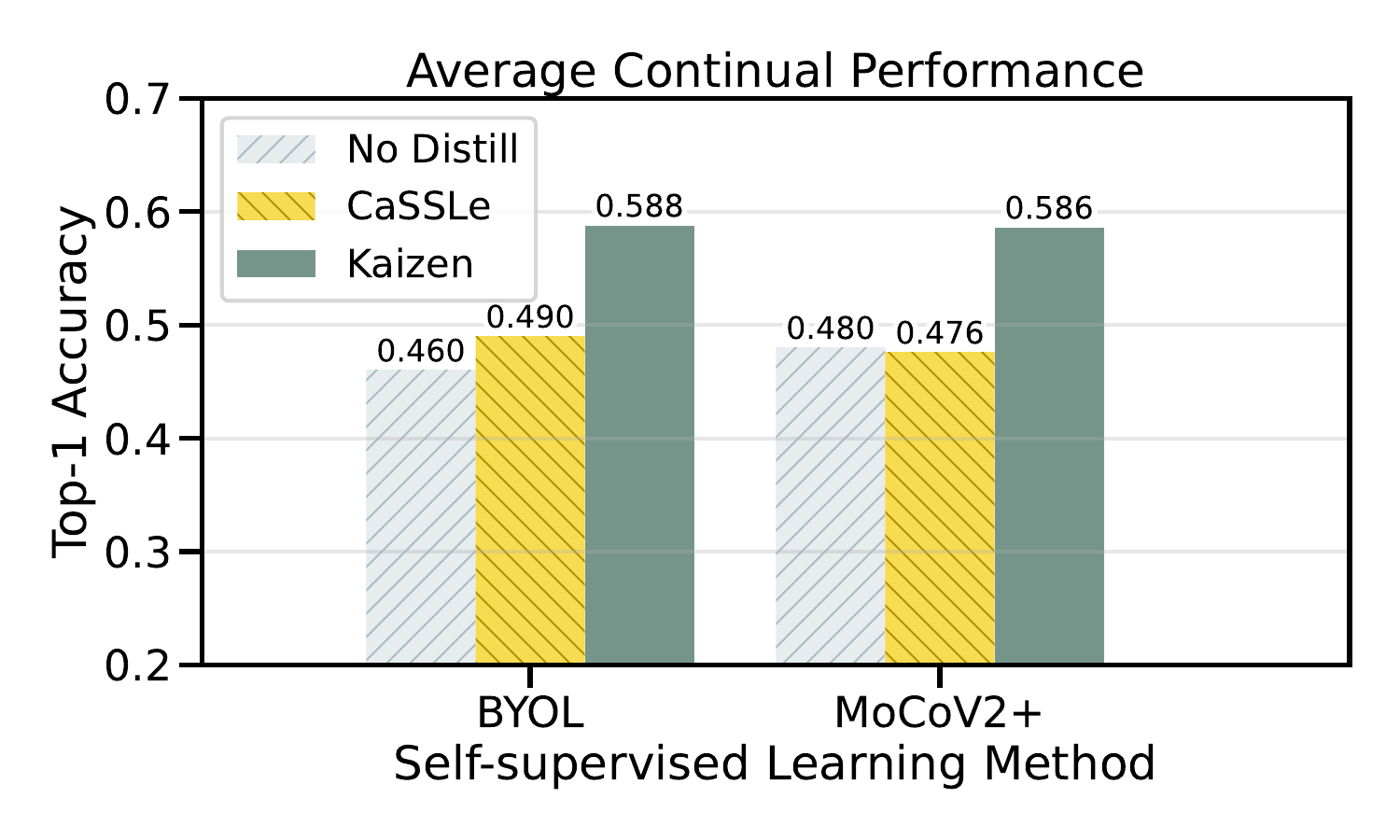}\\
   \vspace{-0.1in}
   \includegraphics[width=0.7\linewidth]{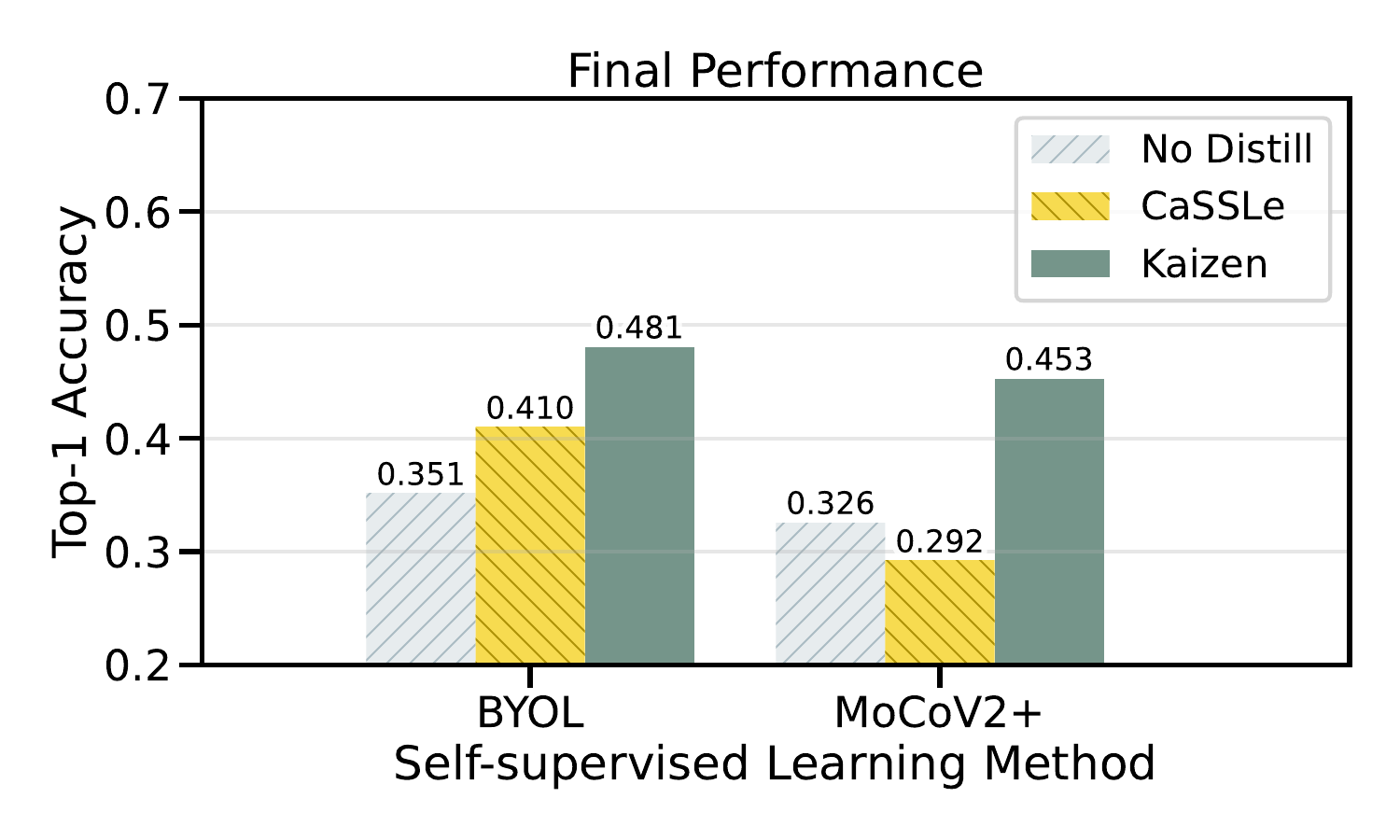}
   
\end{center}
\vspace{-0.2in}
   \caption{Performance comparison between different training methods. Models are trained using different self-supervised learning methods and knowledge distillation strategies on class-incremental WISDM2019. The top figure shows the average performance across the entire continual learning process, while the bottom figure shows the performance in the final evaluation. 
   }
       \vspace{-0.2in}
   \label{fig:general_performance_comparison}
\end{figure}

To address this data scarcity, continual self-supervised learning (CSSL) leverages self-supervision for continual learning. Recent works \cite{fini2022self, tang2023practical} have demonstrated effectiveness in leveraging different sources of data in continual learning, proposing practical solutions with more practical data assumptions. A discussion about related work can be found in Appx.~\ref{related}.

In this study, we explore CSSL techniques for sensor-based human activity recognition. In particular, we adapted CaSSLe \cite{fini2022self} and Kaizen \cite{tang2023practical} to enable self-supervised and semi-supervised continual learning for activity recognition using accelerometer data. Beyond comparing state-of-the-art methods on comprehensive metrics, we investigated adjusting continual fine-tuning's objective importance. Balancing knowledge retention and new task learning with an adaptive weight corresponding to learned/new class ratios proved most effective for performance across tasks. This work demonstrates realistic continual learning assumptions and balancing objectives to achieve the desired performance for mobile sensing. The results highlight the trade-offs in emphasising knowledge retention or new concepts based on use cases.

%% file: sections/3-method.tex
\section{Method}
In this work, we adapted the CaSSLe \cite{fini2022self} and Kaizen \cite{tang2023practical} continual learning frameworks to HAR. This section begins with an overview of these methods and then illustrates the modifications adopted to explore their application to HAR.

\subsection{CaSSLe}
The CaSSLe framework \cite{fini2022self} re-purposed the Siamese/Contrastive learning setup and loss functions for tackling catastrophic forgetting in representation learning. In particular, the framework consists of two main components: new task learning and knowledge distillation.
The \textbf{new task learning (feature extractor)} component follows the conventional Siamese/Contrastive learning setup, like that in BYOL~\cite{grill2020bootstrap} and MoCoV2+~\cite{chen2020improved, he2020momentum}, in which the input signal is transformed into two views using stochastic transformation functions, and the loss function forces the transformed view to have similar representations. The loss term for this is denoted by $\mathcal{L}^{\mathrm{CT}}_{\mathrm{FE}}$.
The \textbf{knowledge distillation (feature extractor)} component mirrors the new task learning component, where the same contrastive loss function is used again, but contrasting the representations retrieved by the feature extractor from the previous task and the current feature extractor instead. An additional predictor (a shallow neural network) is attached to the current feature extractor before contrastive learning. We denote this loss as $\mathcal{L}^{\mathrm{KD}}_{\mathrm{FE}}$.
The CaSSLe framework was shown to be an effective strategy for continual representation learning from a stream of unlabelled data, but it does not propose a specific strategy for training the downstream classifier continually.

\subsection{Kaizen}
The Kaizen framework \cite{tang2023practical} extends CaSSLe by proposing two additional components to handle classifier training, to ensure that a functional classifier is available at any step of the continual learning process.
The \textbf{new task learning (classifier)} component follows conventional supervised learning, in which the classifier is trained using categorical cross-entropy to learn the new classes ($\mathcal{L}^{\mathrm{CT}}_{\mathrm{C}}$).
The \textbf{knowledge distillation (classifier)} component leverages self-distillation to retain knowledge, in which the predictions from the classifier from the previous task are used as pseudo-labels to train the current classifier ($\mathcal{L}^{\mathrm{KD}}_{\mathrm{C}}$). This component does not rely on labelled data and can remain active when only unlabelled data is available.
A small part of the labelled data is retained and replayed, in a similar fashion to other exemplar-based continual learning methods \cite{rebuffi2017icarl, isele2018selective, rolnick2019experience, mittal2021essentials}. This was shown to be a critical component in maintaining classification performance in class-incremental settings \cite{de2021continual}.

\subsection{CSSL for HAR}
\label{subsec:balancing}
As both CaSSLe and Kaizen were proposed for visual representation learning in their original work, we made a few modifications to these frameworks to adapt to HAR.

\subsubsection{Transformation Functions}
Instead of using image transformation functions, we adopted three transformation functions from previous works \cite{har_transformations, multi_self_har, tang2020exploring, tang2021selfhar} that are tailored to sensor time-series: \emph{random 3D rotation}, \emph{random scaling} and \emph{time warping} (see Appx.~\ref{appx:trans}).

\subsubsection{Balancing Learning Objectives}

In addition to following the original formulation of Kaizen, in which the loss function is a sum of all the learning objectives mentioned above 
\[
 \mathcal{L_\mathrm{Kaizen}} =\mathcal{L}^{\mathrm{CT}}_{\mathrm{FE}} + \mathcal{L}^{\mathrm{KD}}_{\mathrm{FE}} + \mathcal{L}^{\mathrm{CT}}_{\mathrm{C}} + \mathcal{L}^{\mathrm{KD}}_{\mathrm{C}}
\]
we hypothesise that the relative importance of the knowledge distillation task compared to learning from new data in classification learning can have a direct impact on the performance of the classifier across time. Therefore, we introduce an importance coefficient $\lambda_{\mathrm{C}}$ to the loss function which allows us to change the weighting of the learning objectives:
\[
 \mathcal{L_\mathrm{Kaizen(adaptive)}} =(\mathcal{L}^{\mathrm{CT}}_{\mathrm{FE}} + \mathcal{L}^{\mathrm{KD}}_{\mathrm{FE}}) + (\mathcal{L}^{\mathrm{CT}}_{\mathrm{C}} + \lambda_{\mathrm{C}} \mathcal{L}^{\mathrm{KD}}_{\mathrm{C}})
\]
The effects of this importance coefficient are explored in our evaluation.

%% file: sections/4-evaluation.tex
\section{Evaluation and Results}
In our work, we compare different state-of-the-art CSSL methods in different setups (see Appx.~\ref{appx:setup} for details regarding our evaluation setup).

\subsubsection{Performance Comparison against CSSL}

Here, we focus on the overall performance of Kaizen, CaSSLe and the \emph{No distill} setup. In Fig.~\ref{fig:general_performance_comparison}, we present a comparative analysis of Continual Accuracy and Final Accuracy across various methods, each employing different (SSL) models for the feature extractor. This comparison is specifically conducted on the WISDM2019 dataset, which has been divided into six equal-sized tasks. We observe that Kaizen consistently surpasses the two baseline models across all evaluation metrics and SSL methods. Specifically, using BYOL, Kaizen achieved the highest continual accuracy (0.588) and final accuracy (0.481), exceeding CaSSLe by 0.098 and 0.071, respectively. Notably, data availability is standardized across methods, with each having access to current task data and 1\% replay data from previous tasks. Kaizen's superiority is attributed to integrating knowledge distillation and fine-tuning into the pipeline, rather than solely training the classifier at the end.

\subsubsection{Performance Variation across Time}

\begin{figure}
    \centering
    \includegraphics[width=0.49 \linewidth ]{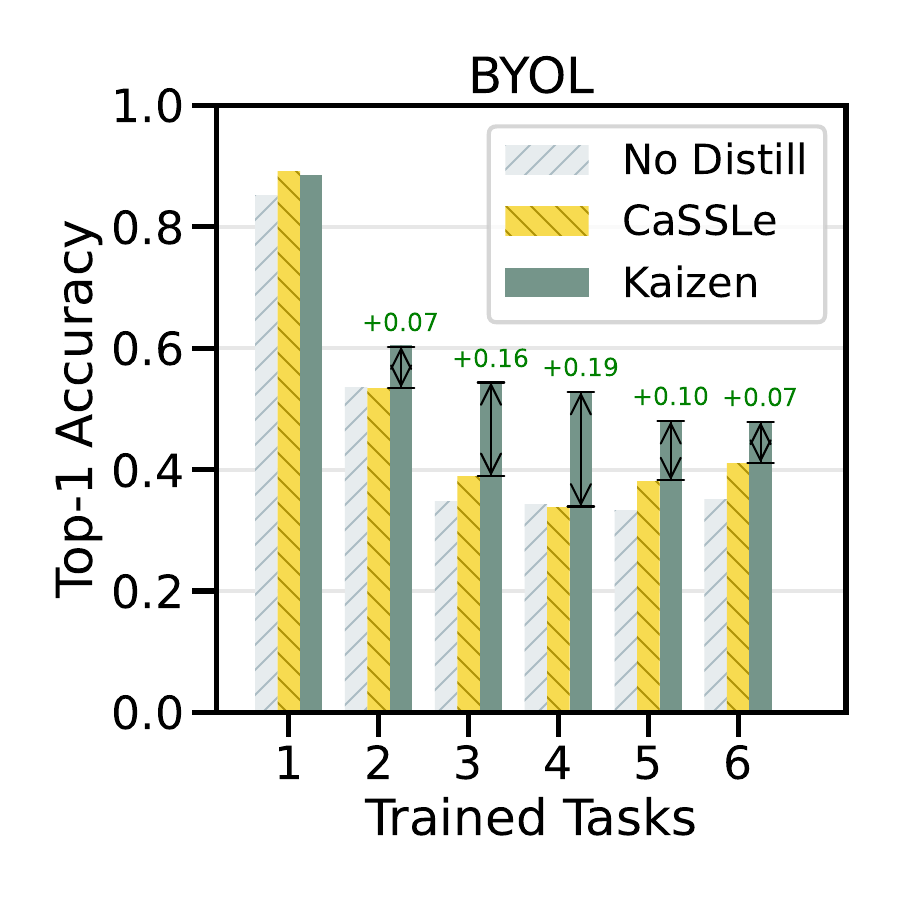}
    \includegraphics[width=0.49 \linewidth ]{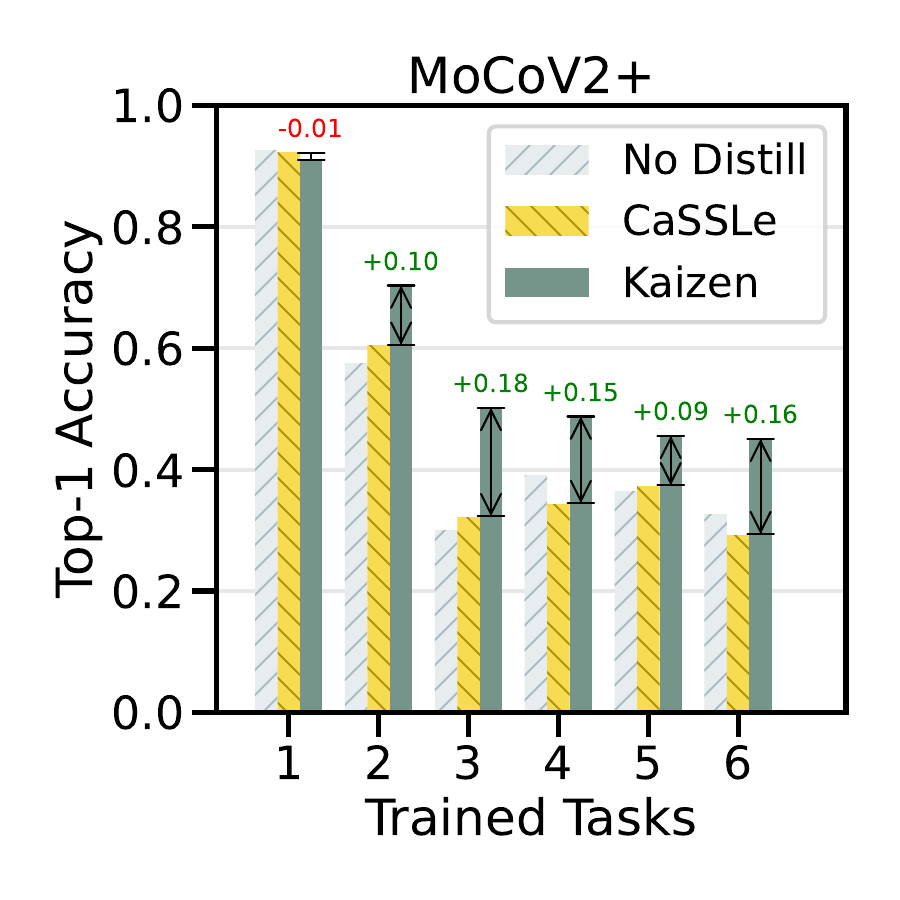}
    \vspace{-0.2in}
    \caption{Average performance over tasks on WISDM2019. Comparison is drawn between training methods using different SSL algorithms across different tasks.}
        \vspace{-0.1in}
    \label{fig:performance_across_time_cifar}
\end{figure}

In this evaluation, we examine the performance of different methods at different stages of the continual learning process. Fig.~\ref{fig:performance_across_time_cifar} shows the average accuracy over trained tasks of Kaizen, CaSSLe and \emph{No distill} after training on each task. We find that in general, all methods show lower performance after training on more tasks, partially because the classification problem becomes more difficult as the number of classes increases, as well as catastrophic forgetting. Echoing the results of the previous evaluation, Kaizen maintains a higher level of accuracy overall, even though all methods start from similar performance on the first task.

\subsubsection{Per-task Performance Breakdown} \label{subsection:per_task} 
\begin{figure}[t]
    \centering
    \includegraphics[width=0.32 \linewidth ]{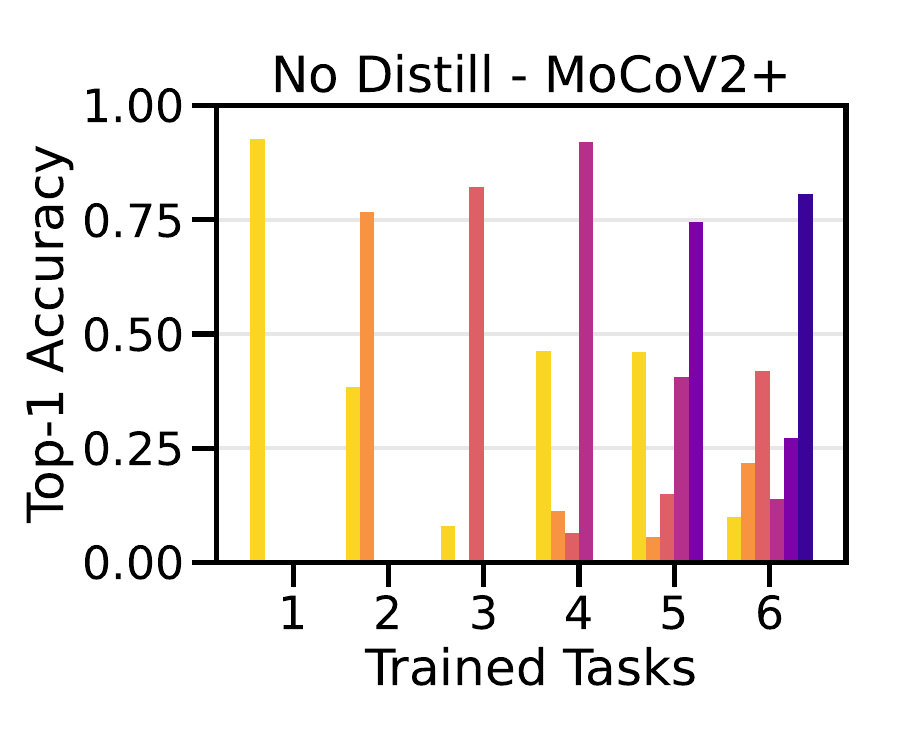}
    \includegraphics[width=0.32 \linewidth ]{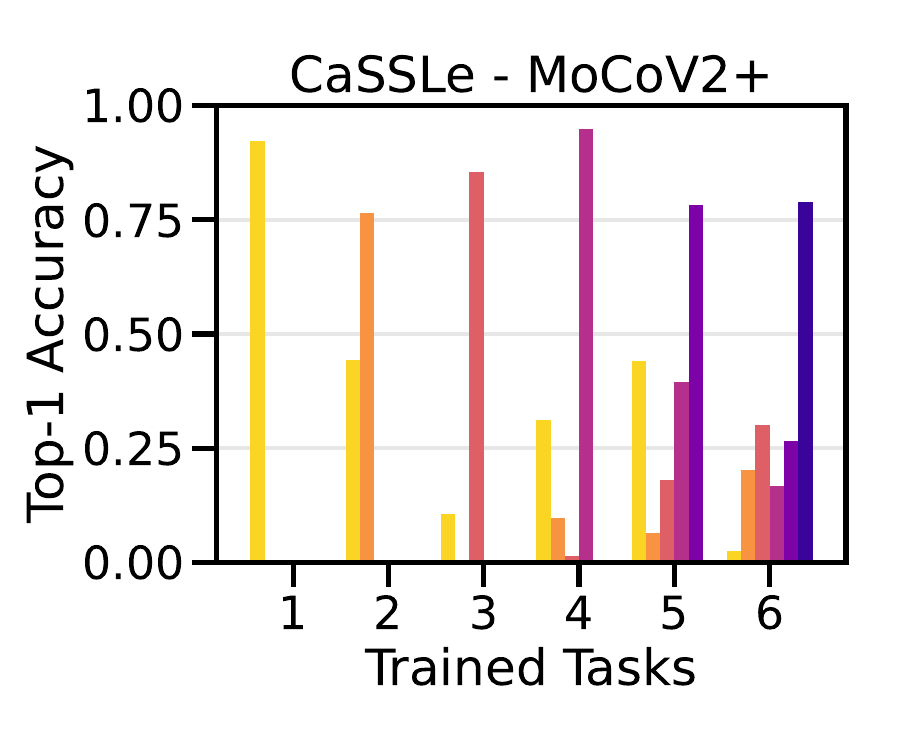}
    \includegraphics[width=0.32 \linewidth ]{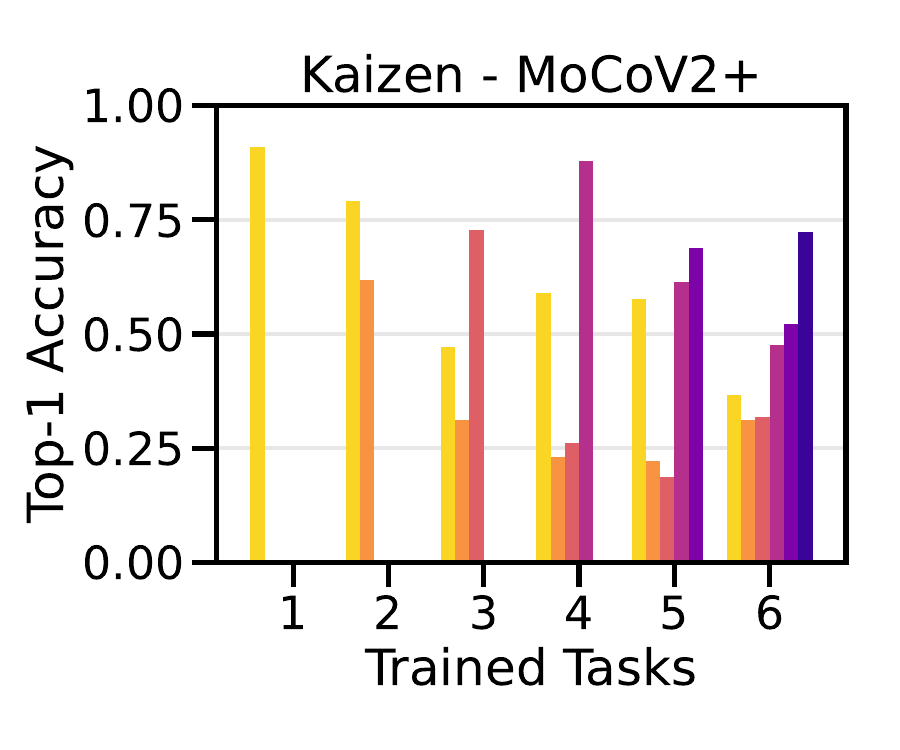} \\
    \vspace{-0.05in}
    
    \includegraphics[width=0.32 \linewidth ]{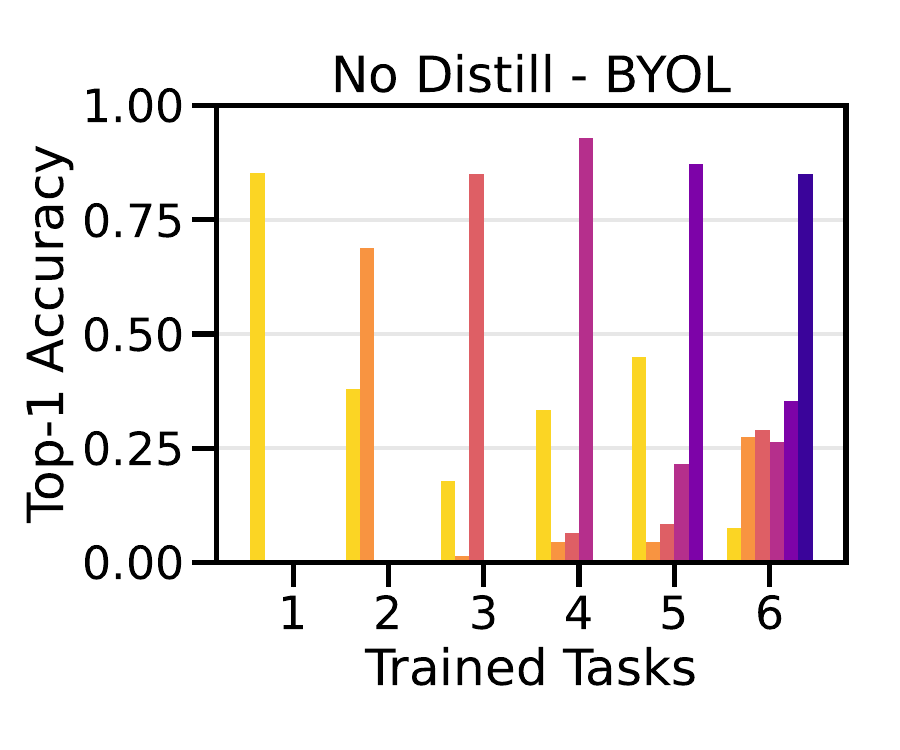}
    \includegraphics[width=0.32 \linewidth ]{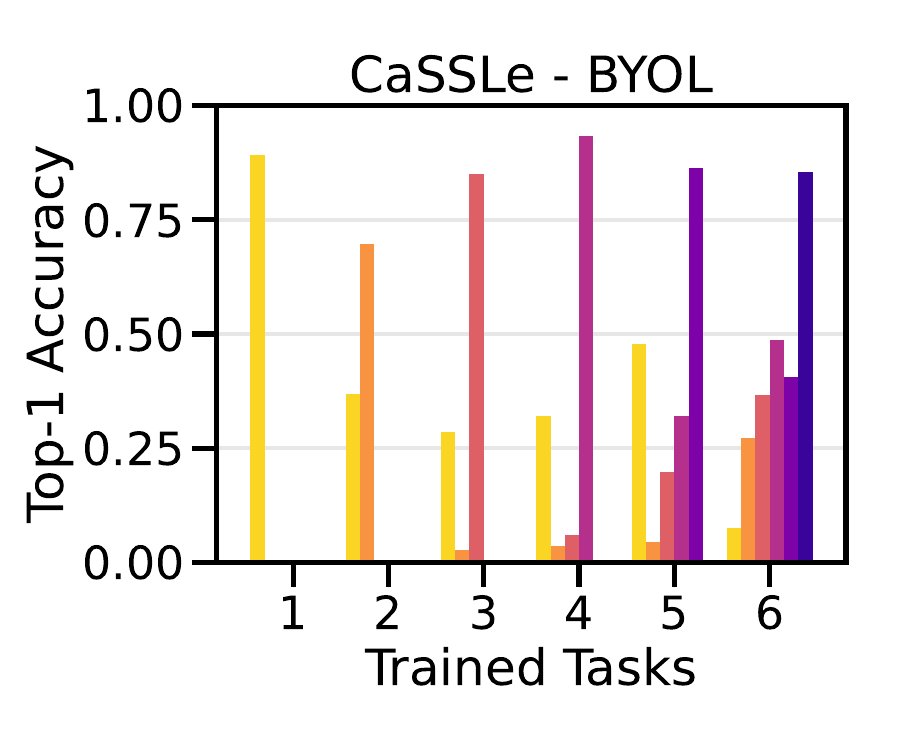}
    \includegraphics[width=0.32 \linewidth ]{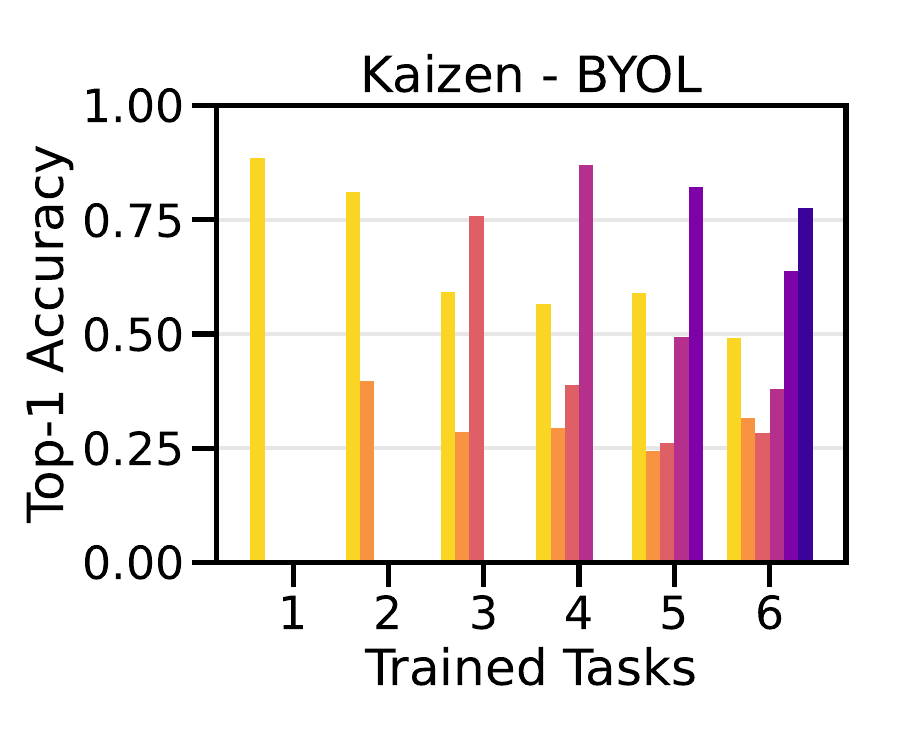} \\
    \vspace{-0.05in}
    \includegraphics[width=0.65 \linewidth ]{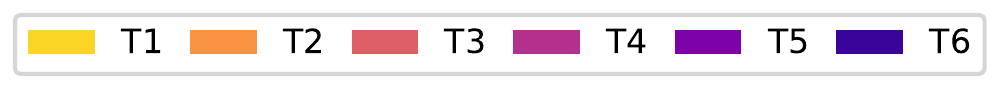}
    \vspace{-0.1in}
    
    \caption{Detailed breakdown of performance over tasks on WISDM2019. Fine-grained accuracy for every task is shown.}
        
    \label{fig:performance_per_task}
\end{figure}

\begin{table}[t!]
\caption{SSL and baseline performance comparison. We evaluate three continual learning strategies across four metrics on WISDM2019. Kaizen with BYOL outperforms in most metrics. The best performance for a SSL method is bolded, and that across SSL methods is underlined. (FA: Final Accuracy, CA: Continual Accuracy, F: Forgetting, FT: Forward Transfer).}

\begin{center}
\small
\begin{tabularx}{\linewidth}{c!{\vrule width 1.5pt}c!{\vrule width 1.5pt}*{4}c}

\textbf{SSL}        &     \textbf{Baseline}     & \textbf{FA} $\uparrow$ & \textbf{CA} $\uparrow$ & \textbf{F} $\downarrow$ & \textbf{FT} $\uparrow$ \\
\hline \hline
\multirow{3}{*}{BYOL}    & No Distill & 0.351          & 0.460              & 0.587          & 0.006            \\
                         & CaSSLe     & 0.410          & 0.490              & 0.527          & \textbf{0.008}   \\
                         & Kaizen       & \textbf{\underline{0.481}} & \textbf{\underline{0.588}}     & \textbf{\underline{0.408}} & -0.107           \\ \hline
\multirow{3}{*}{MoCoV2+}  & No Distill & 0.326          & 0.480              & 0.606          & 0.007            \\
                         & CaSSLe     & 0.292          & 0.476              & 0.662          & \textbf{\underline{0.023}}   \\
                         & Kaizen       & \textbf{0.453} & \textbf{0.586}     & \textbf{0.401} & -0.077           \\
\end{tabularx}
    \vspace{-0.2in}
\end{center}

\label{tab:ssl_baseline}
\end{table}
Kaizen prioritises knowledge retention over the acquisition of new task proficiency. This is evidenced through an analysis of per-task performance at sequential stages, as depicted in Fig. \ref{fig:performance_per_task}. Here, the performance metrics of the models on individual tasks are evaluated through the continual learning process, utilizing MoCoV2+ and BYOL as the self-supervised learning method. Notably, Kaizen demonstrates more graceful knowledge forgetting in comparison to CaSSLe, where there is a gradual decline in performance on prior tasks. While CaSSLe excels in adapting to new tasks, it shows a significant and rapid deterioration in performance on previously learned tasks. Kaizen demonstrated a more balanced trade-off between mastering new tasks and retaining knowledge from past tasks.

\subsubsection{Comprehensive Evaluation of Continual Learning and SSL Methods}

Table \ref{tab:ssl_baseline} presents the performance of Kaizen, CaSSLe and \emph{No Distill} on different metrics. These results show that Kaizen achieved the best performance in terms of the final and continual accuracy. 
As expected, the absence of distillation from the \emph{No Distill} approach hurts the accuracy of the overall continual learning framework as it is unable to maintain knowledge with each additional task. This can also be seen in the Forgetting metric where \emph{No Distill} is 0.179 higher than Kaizen and 0.06 higher than CaSSLe when BYOL is used as the SSL method. BYOL in general performs better as the contrastive learning and knowledge retention mechanism compared to MoCoV2+. On the other hand, as shown in the results above, CaSSLe and \emph{No Distill}, show a slight positive forward transfer in some cases, while Kaizen suffers from a negative forward transfer. This can be explained by the tendencies of CaSSLe and \emph{No Distill} prioritising learning from new data and outperforming specialised models.

\subsubsection{Influence of the Importance Coefficient for Continual Fine-tuning}

As introduced in Section~\ref{subsec:balancing}, the importance coefficient $\lambda_{\mathrm{C}}$ specifies the relative importance of the knowledge distillation task compared to learning from new data. As such, it can have a direct impact on the performance of the classifier across time: a higher importance coefficient forces the model to focus on knowledge retention, while a lower coefficient shifts this focus to new task learning.

\begin{figure}[t]
\begin{center}
   \includegraphics[width=0.49\linewidth]{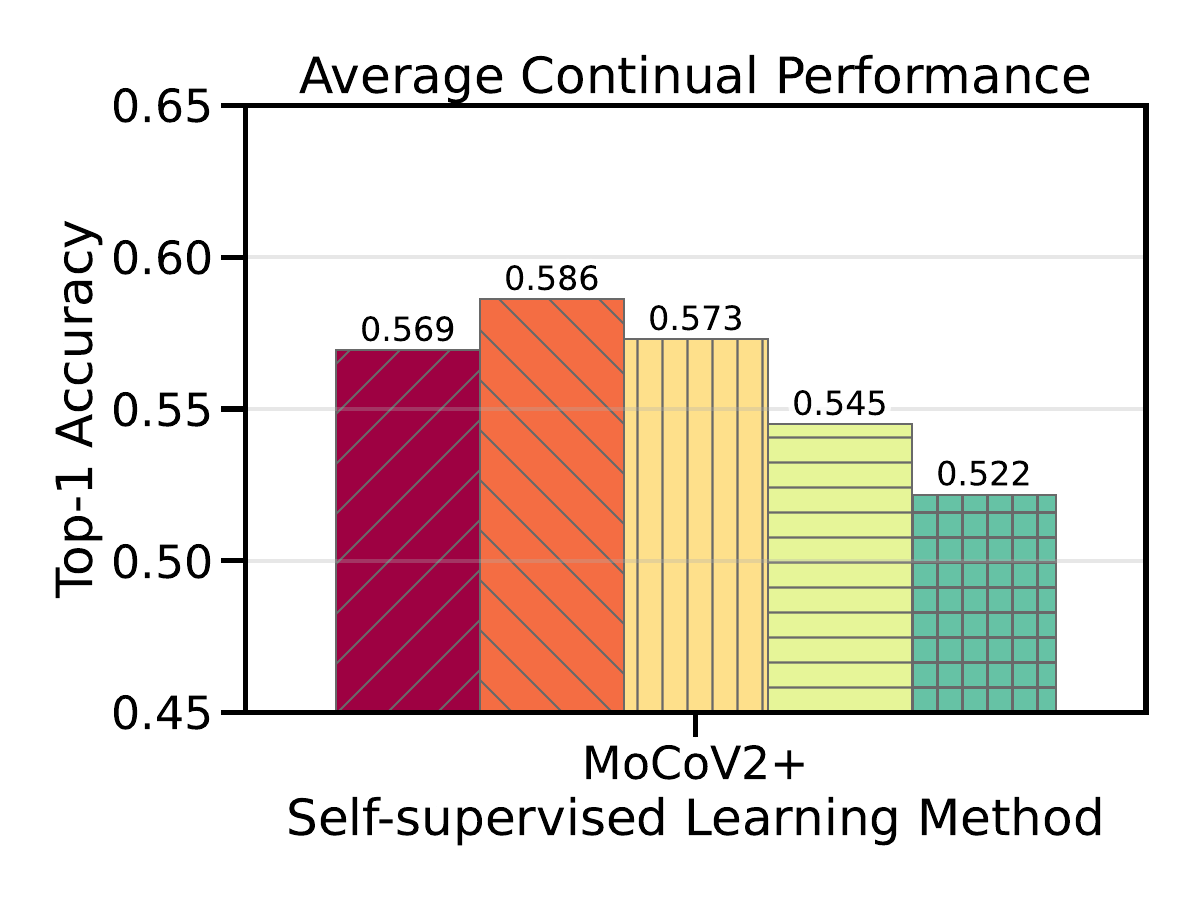}
   \includegraphics[width=0.49\linewidth]{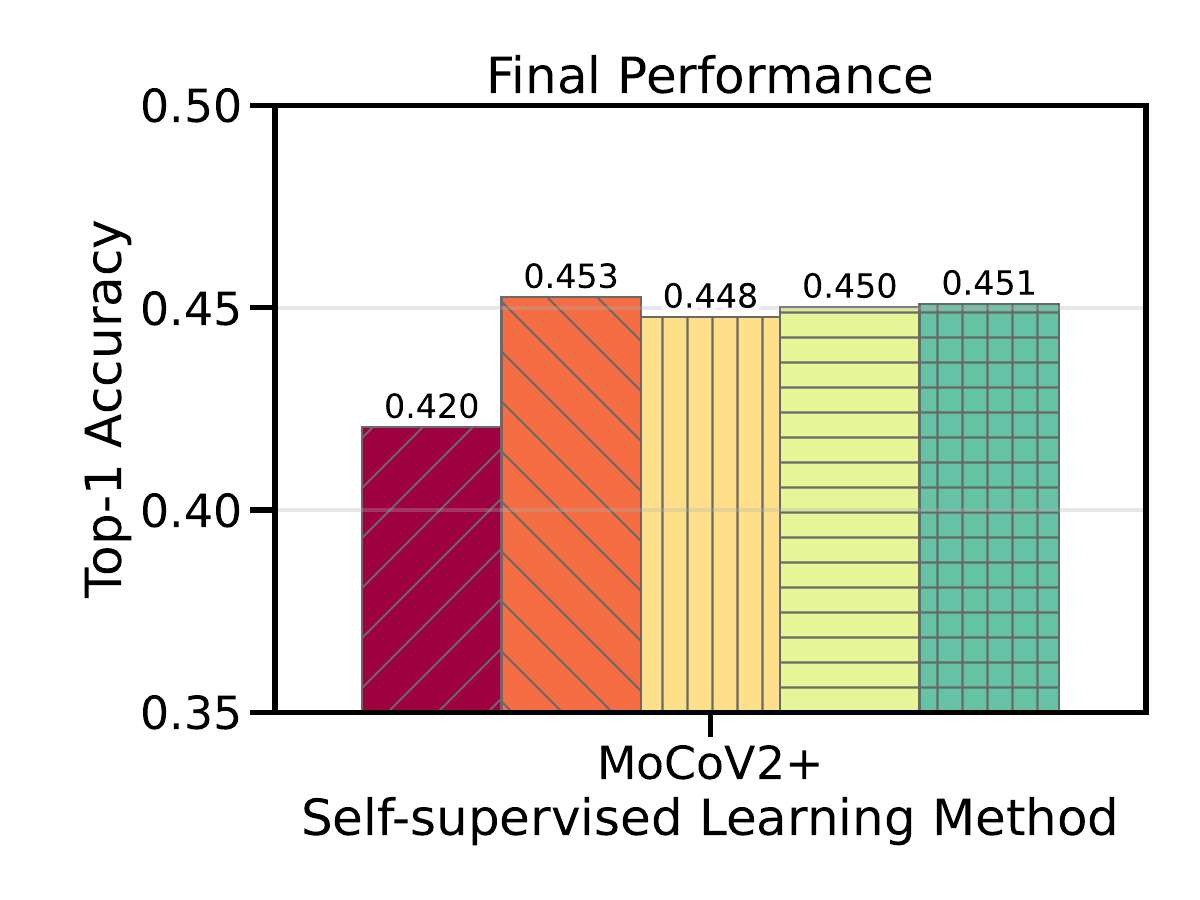} \\
   \vspace{-0.05in}
   \includegraphics[width=0.8\linewidth]{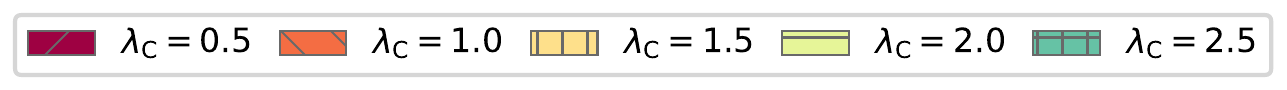}
\end{center}
\vspace{-0.15in}
   \caption{Effects of constant importance coefficients. The plot compares the aggregate performance metrics of models trained with the importance coefficient for the classifier set to different constant values.}
    \vspace{-0.2in}
   \label{fig:kaizen_general_performance_constant_lamb}
\end{figure}

\begin{figure}[ht]
    \centering
    \includegraphics[width=0.32 \linewidth ]{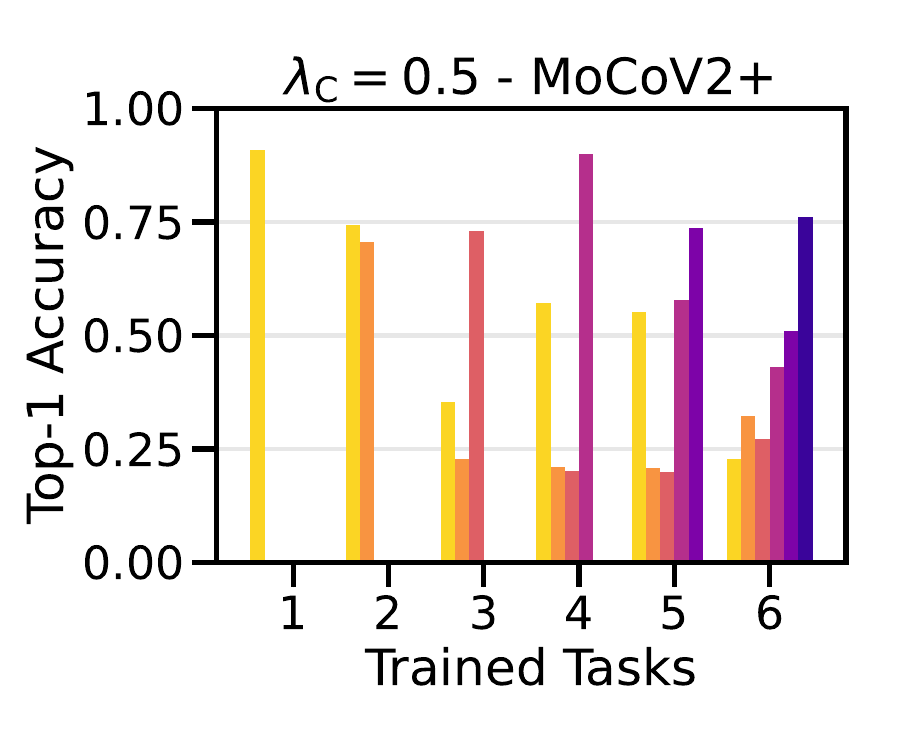}
    \includegraphics[width=0.32 \linewidth ]{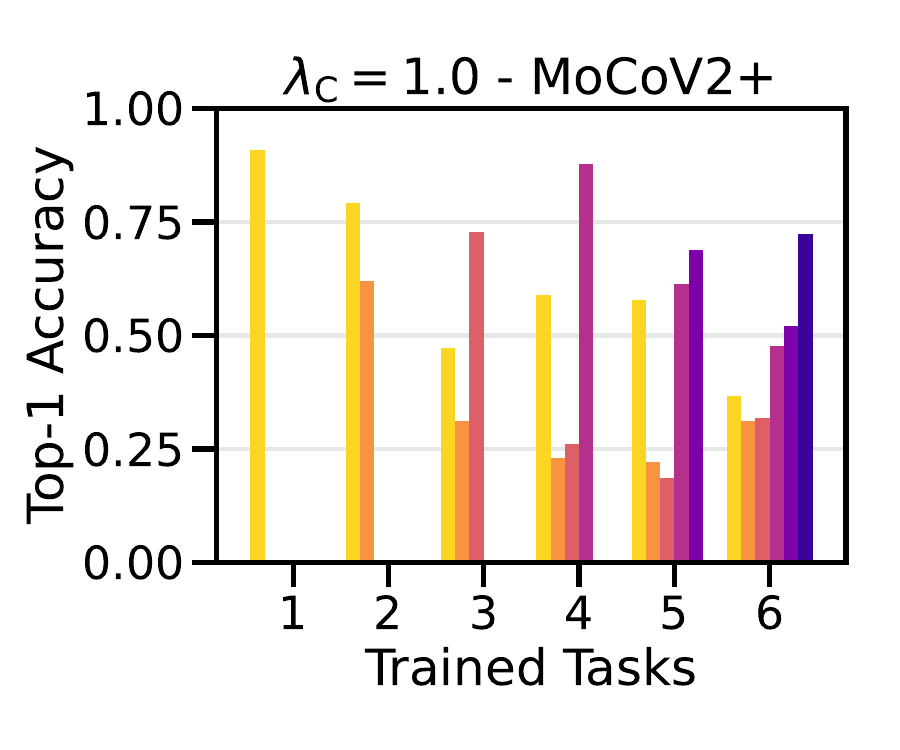}
    \includegraphics[width=0.32 \linewidth ]{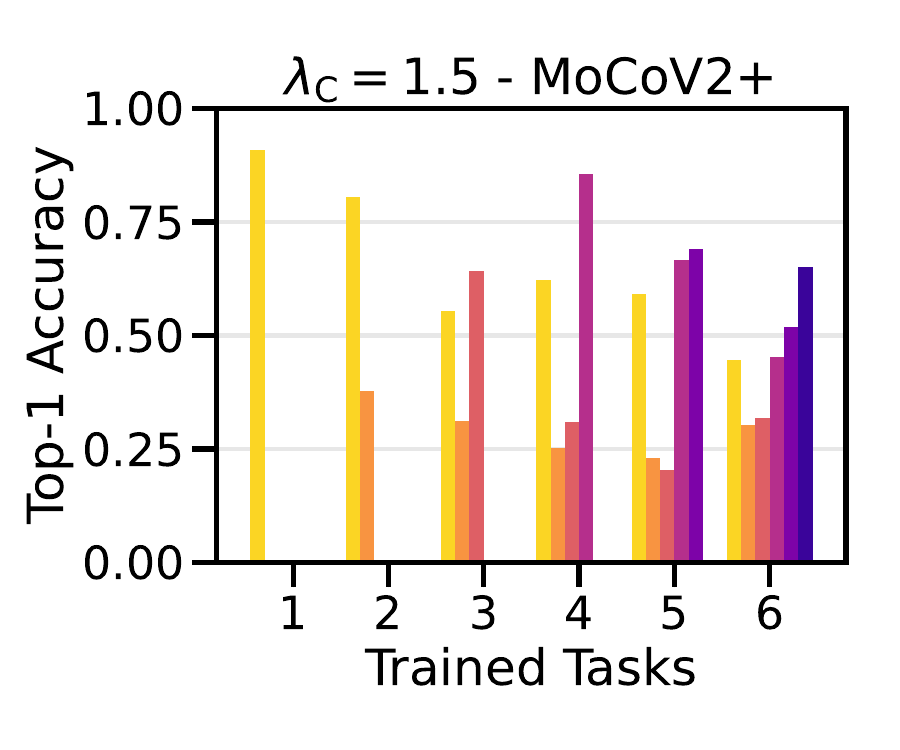} \\
    \vspace{-0.05in}
    \includegraphics[width=0.32 \linewidth ]{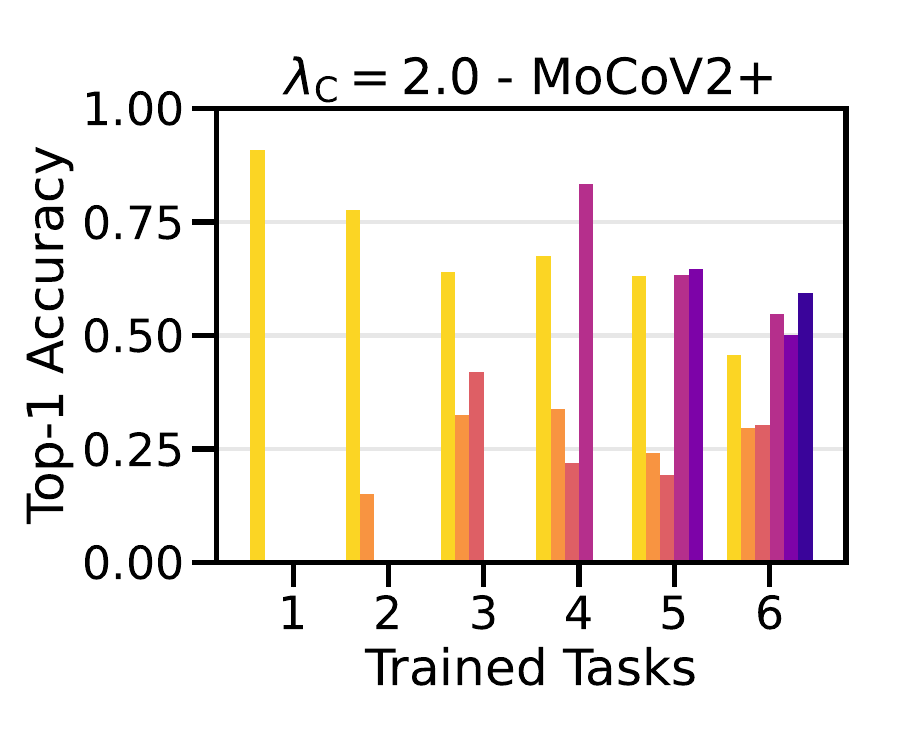}
    \includegraphics[width=0.32 \linewidth ]{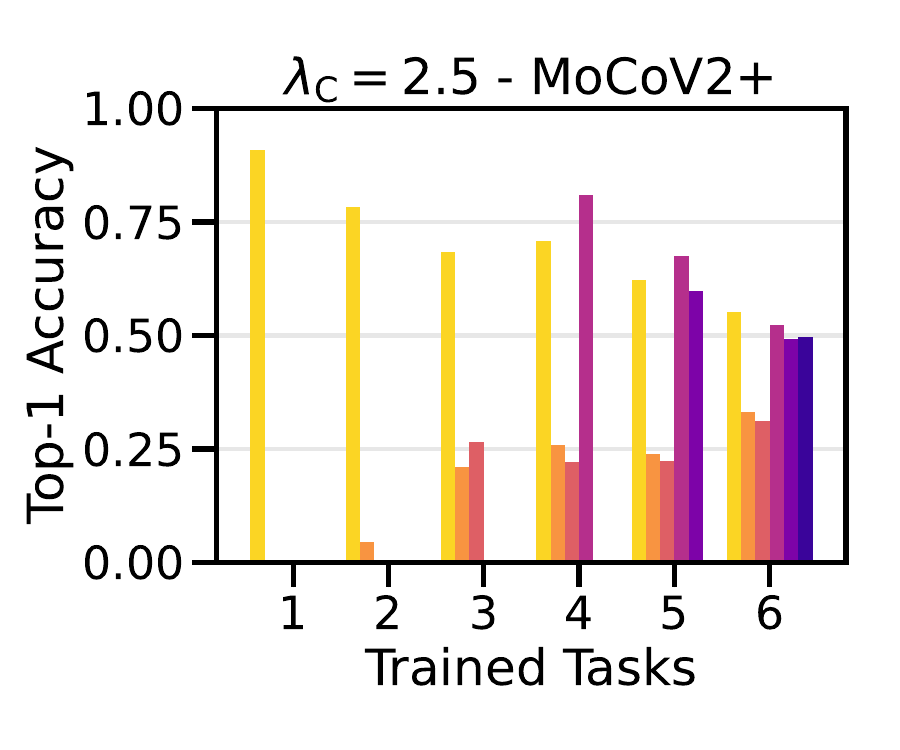} \\
    \vspace{-0.05in}
    \includegraphics[width=0.65 \linewidth ]{figures_new/Part_1/F4-WISDM2019-6Tasks-v2_legend.pdf}
    \vspace{-0.1in}
    \caption{Detailed breakdown of performance over tasks with constant importance coefficients. Fine-grained accuracy is shown for every additional task with different importance coefficients.}
    \label{fig:kaizen_performance_per_task_constant_lamb}
\end{figure}

We conducted an additional set of experiments exploring the effect of this hyperparameter on the performance of the model across time while all other hyperparameters remain unchanged for Kaizen. Fig. \ref{fig:kaizen_general_performance_constant_lamb} and \ref{fig:kaizen_performance_per_task_constant_lamb} illustrate the changes in performance when we set the value of $\lambda_{\mathrm{C}}$ to 0.5, 1.0, 1.5, 2.0, and 2.5 (additional visualisation can be found in Appx.~\ref{subsection:additional_results}). From Fig.~\ref{fig:kaizen_performance_per_task_constant_lamb}, we can see that with lower values of $\lambda_{\mathrm{C}}$, the model tends to have higher accuracy in new tasks right after they have been trained on them, which is similar to that of CaSSLe. Higher values of $\lambda_{\mathrm{C}}$, on the other hand, force the model to retain knowledge on older tasks, which matches our understanding of this hyperparameter.

We also observe a trend here: even though higher values of $\lambda_{\mathrm{C}}$ forces the model to retain more knowledge (see Fig.~\ref{fig:kaizen_performance_per_task_constant_lamb}, where the performance on task 1, denoted by T1, remains high throughout all steps for $\lambda_{\mathrm{C}}=2.5$), the overall performance of the model ends up lower in the long run (see Fig.~\ref{fig:kaizen_general_performance_constant_lamb}). This can be counter-intuitive, but explainable by inspecting $\lambda_{\mathrm{C}}$: a high value of $\lambda_{\mathrm{C}}$ sacrifices new task learning for knowledge retention, and this effect \emph{compounds} over time. Performance on the initial task (task 1) is kept at the highest priority when using a high $\lambda_{\mathrm{C}}$, while all other tasks suffer as the model is being trained. This is reflected in Fig.~\ref{fig:kaizen_performance_per_task_constant_lamb}, where the performance on tasks 2, 3, 4, 5, and 6 starts off with a much lower value when using $\lambda_{\mathrm{C}}=2.5$ compared to $\lambda_{\mathrm{C}}=0.5$. Since the goal of continual learning is to maintain high performance on \emph{all} trained tasks over time, such a trade-off can hurt the overall performance.

We hypothesise that a \emph{progressive} importance coefficient can allow the model to balance knowledge retention and new task learning better than a constant value: as the model gets trained on more classes and tasks, the importance of knowledge retention should increase, and it should be proportional to the number of classes already learned compared to the number of new classes to be learned. To test this hypothesis, we designed a set of experiments where the importance coefficient is increased by a constant value after each task. We denote the setting where $\lambda_{\mathrm{C}}$ is set to $a$ initially, and increased by $b$ after each task as $\lambda_{\mathrm{C}}=a \oplus b$. That is, the importance coefficient at task $t$ is given by $\lambda_{\mathrm{C}}(t) = a + b \times (t - 1)$.
For example, the setting $\lambda_{\mathrm{C}}=1.0 \oplus 0.5$ will see values of $\lambda_{\mathrm{C}}$ being set to 1.0, 1.5, 2.0, 2.5, 3.0, and 3.5 for tasks 1, 2, 3, 4, 5, and 6 respectively, while the setting $\lambda_{\mathrm{C}}=1.0 \oplus 0.0$ is identical to that of a constant $\lambda_{\mathrm{C}}=1.0$.

\begin{figure}[t]
\begin{center}
   \includegraphics[width=0.49\linewidth]{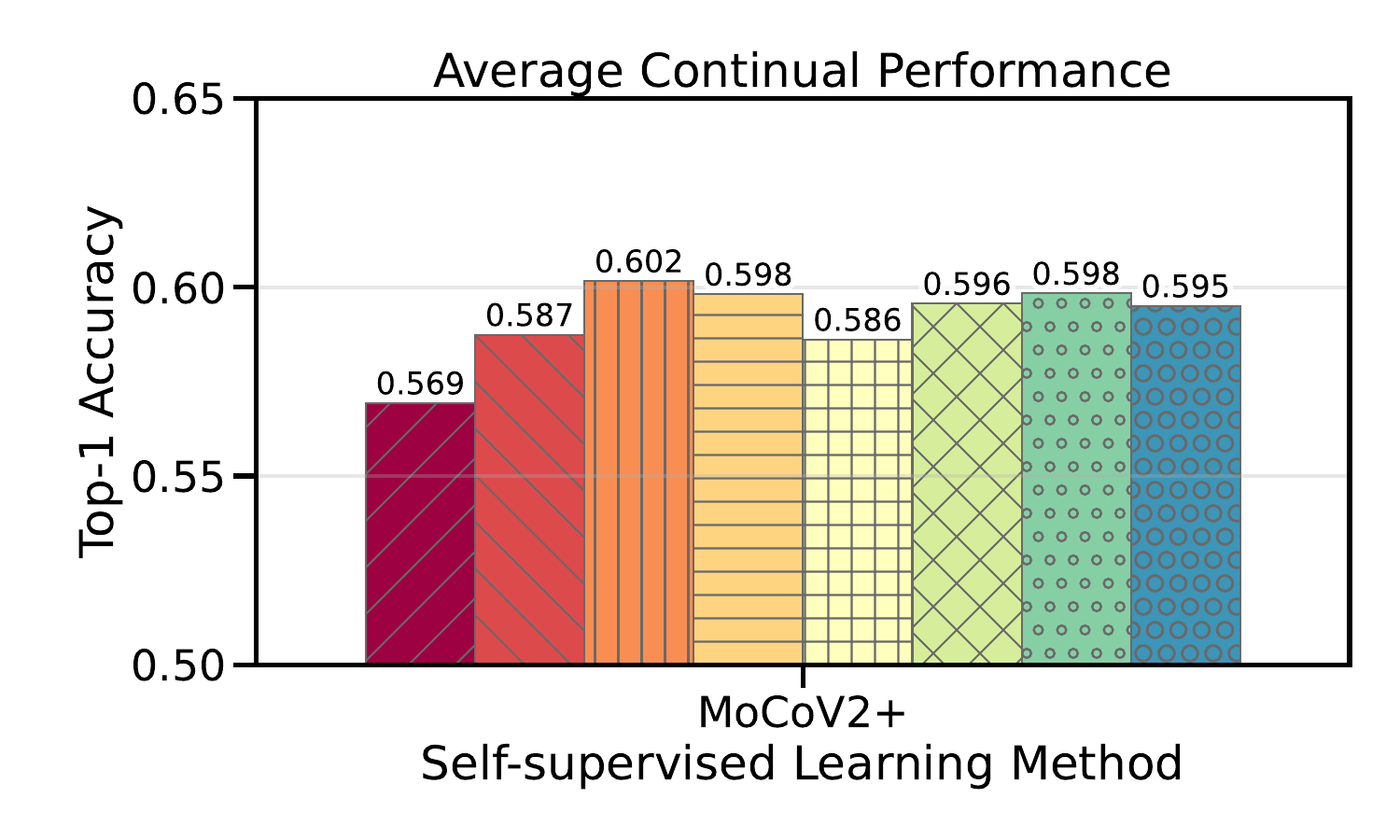}
   \includegraphics[width=0.49\linewidth]{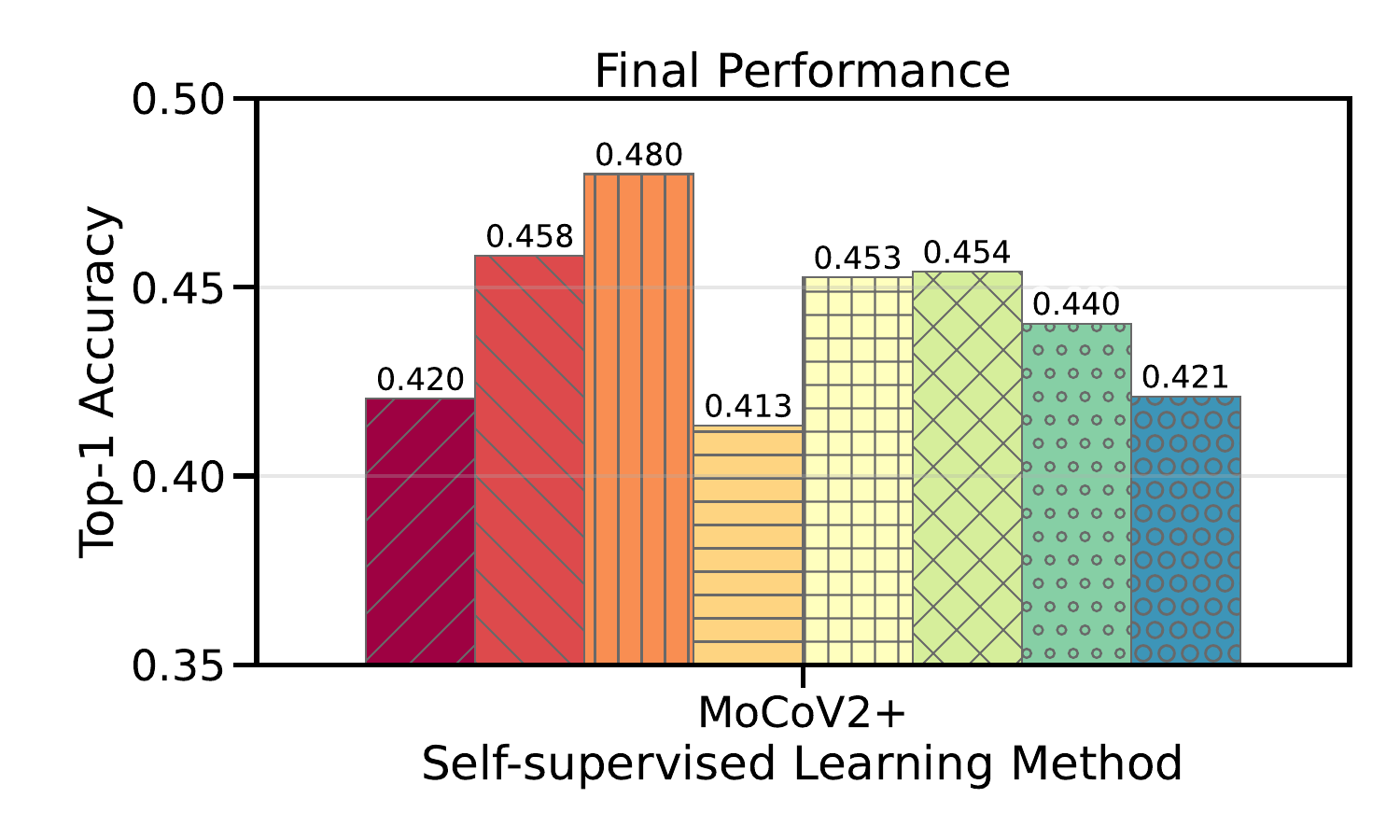} \\
   \vspace{-0.05in}
   \includegraphics[width=0.9\linewidth]{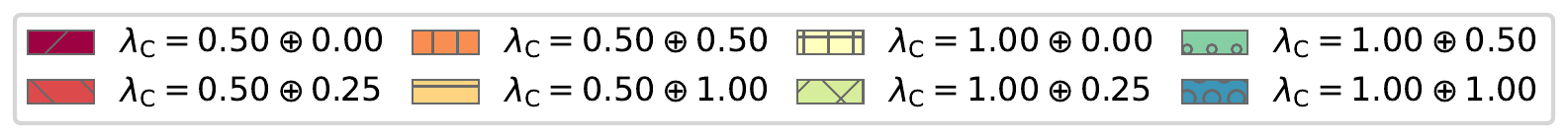}
\end{center}
\vspace{-0.15in}
   \caption{Effects of progressive importance coefficients. The plot compares the aggregate performance metrics of models trained with the importance coefficient for the classifier set to different progressive values.}
   \label{fig:kaizen_general_performance_progressive_lamb}
       \vspace{-0.2in}
\end{figure}

\begin{figure}[ht]
    \centering
    \includegraphics[width=0.24 \linewidth ]{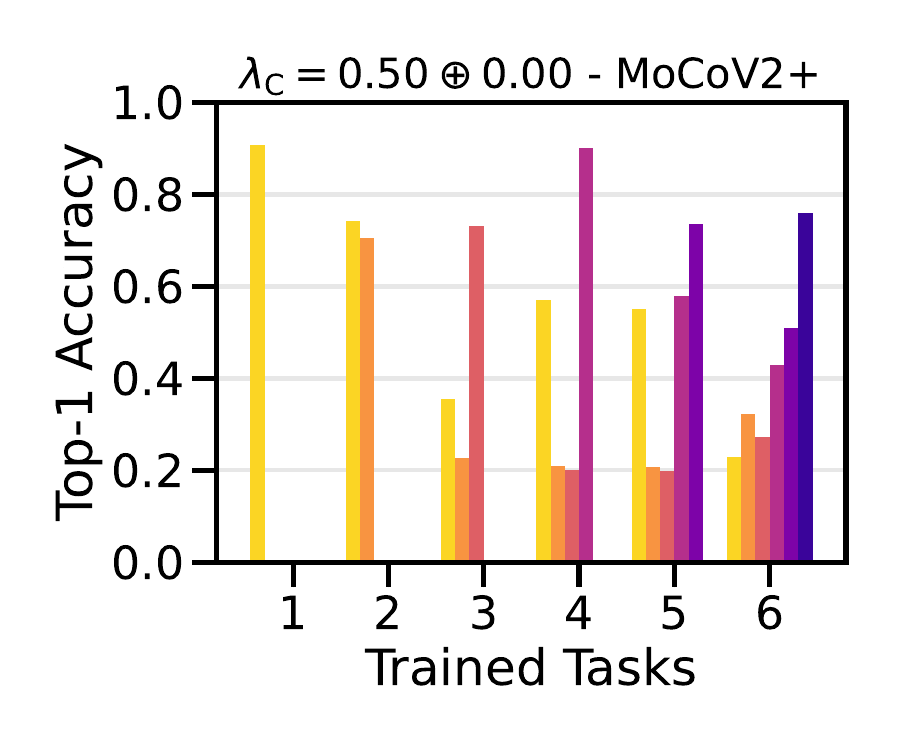}
    \includegraphics[width=0.24 \linewidth ]{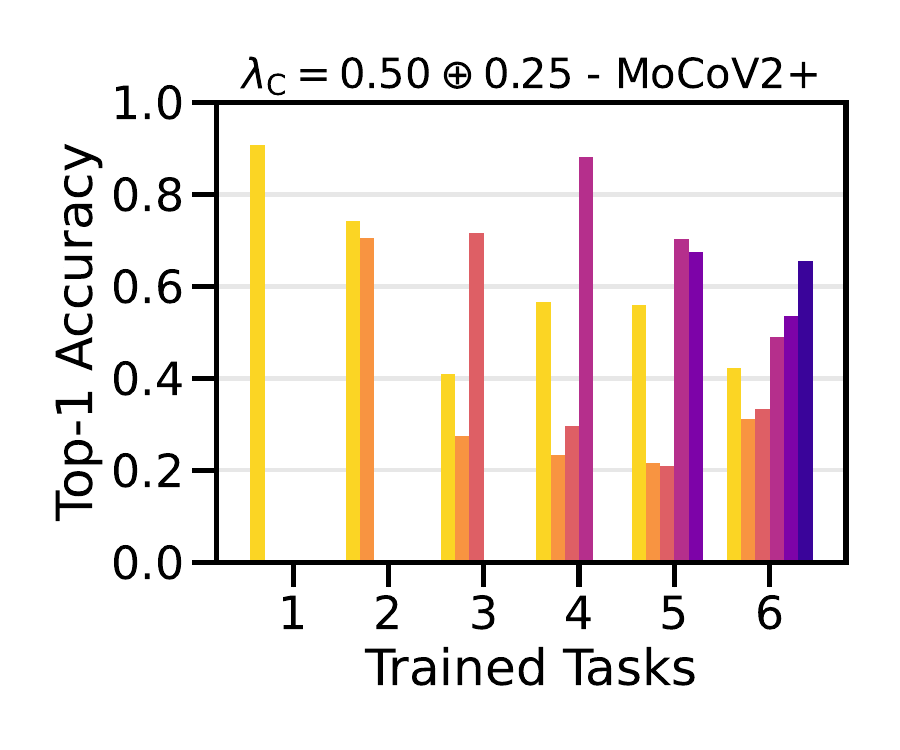}
    \includegraphics[width=0.24 \linewidth ]{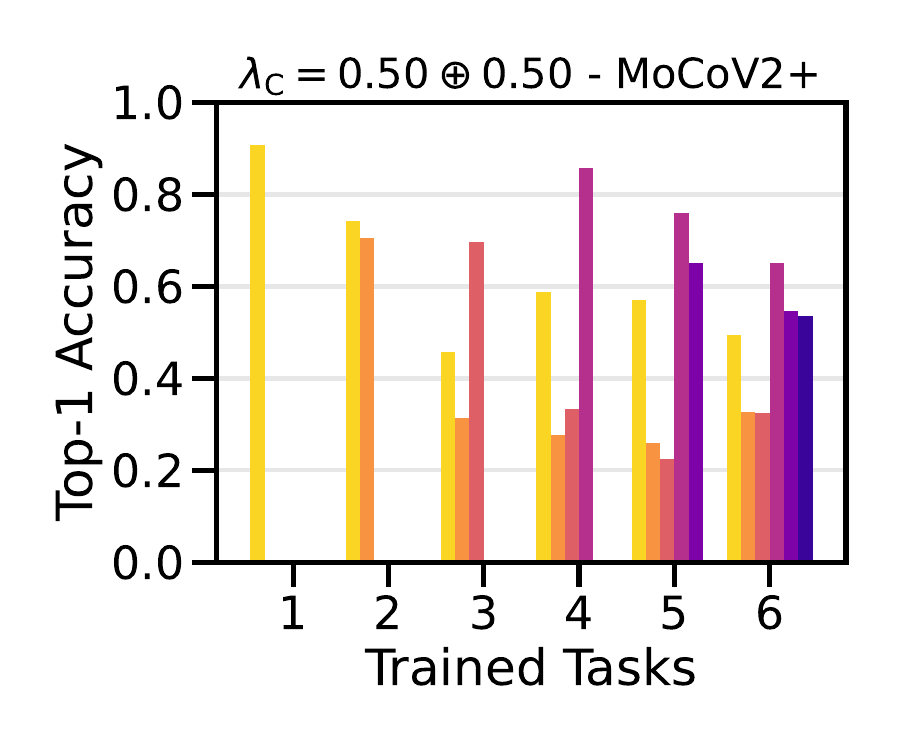}
    \includegraphics[width=0.24 \linewidth ]{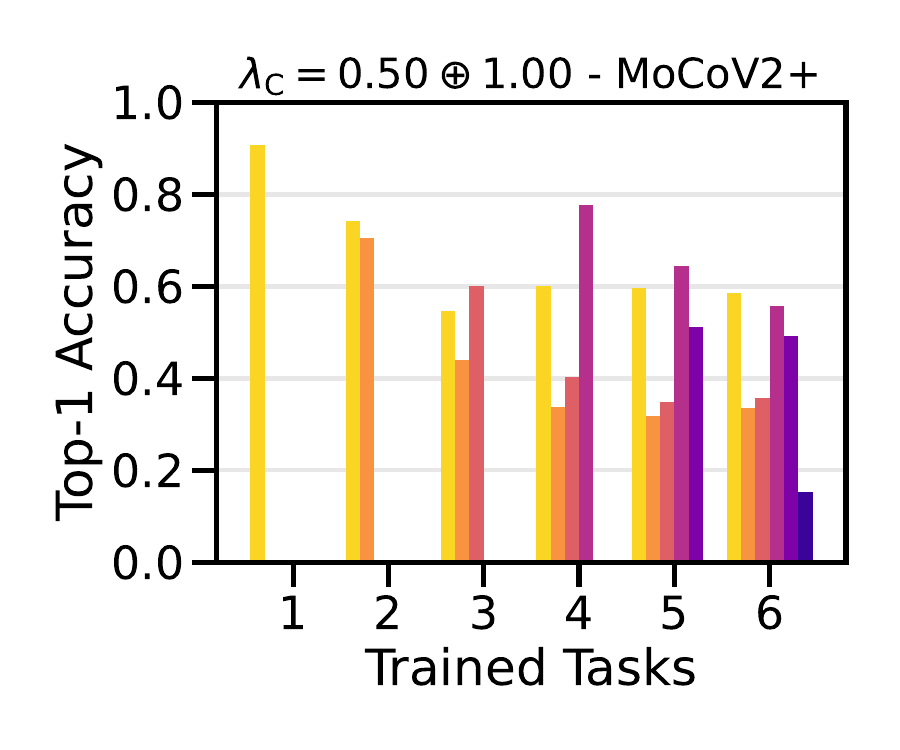} \\
    \vspace{-0.05in}
    \includegraphics[width=0.24 \linewidth ]{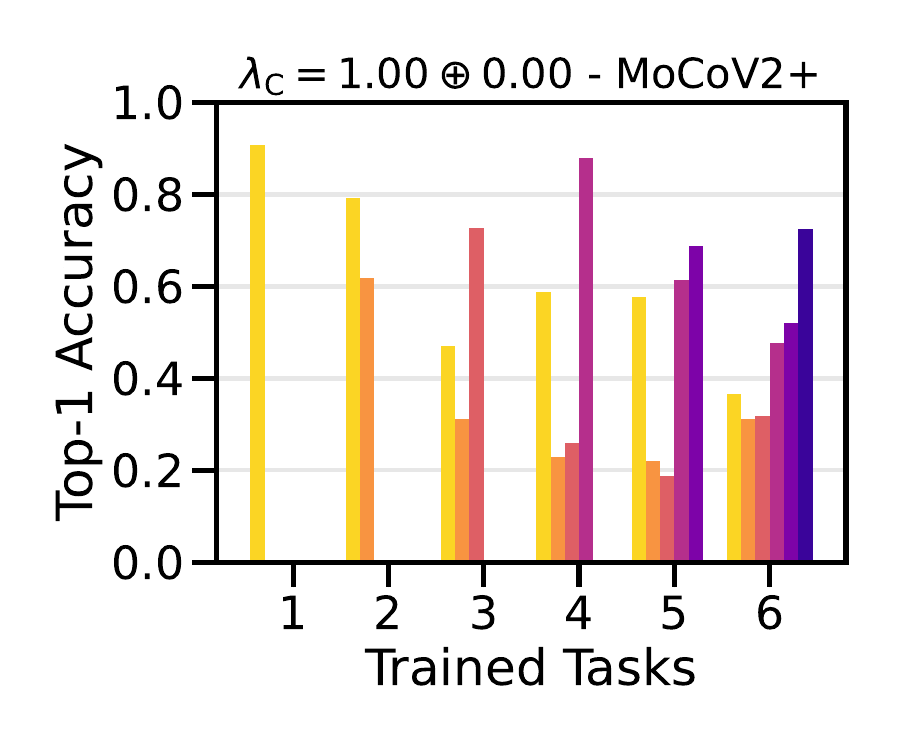}
    \includegraphics[width=0.24 \linewidth ]{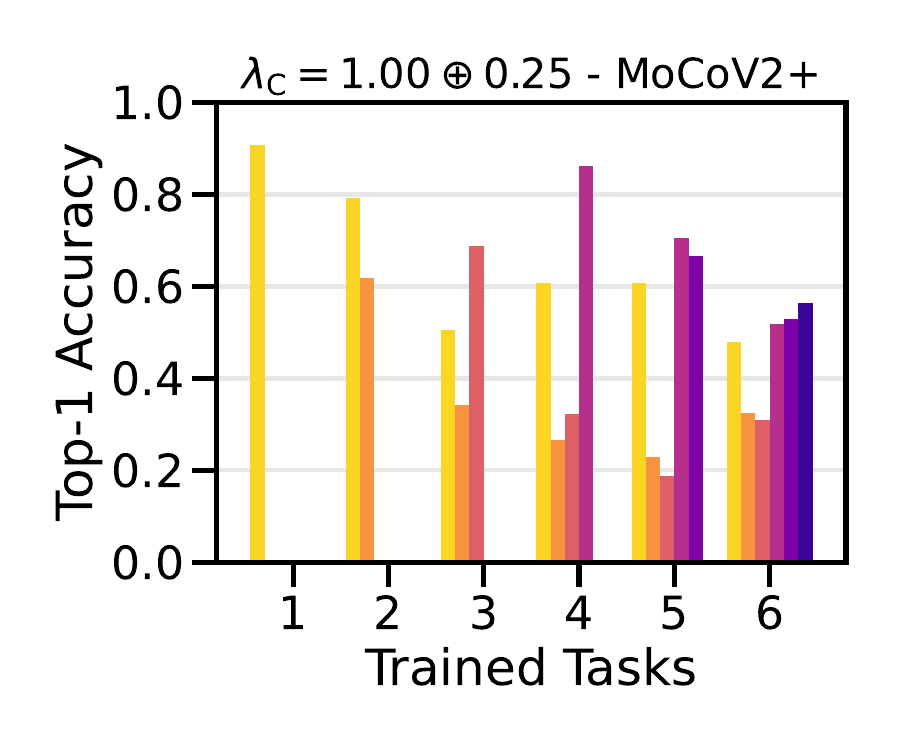}
    \includegraphics[width=0.24 \linewidth ]{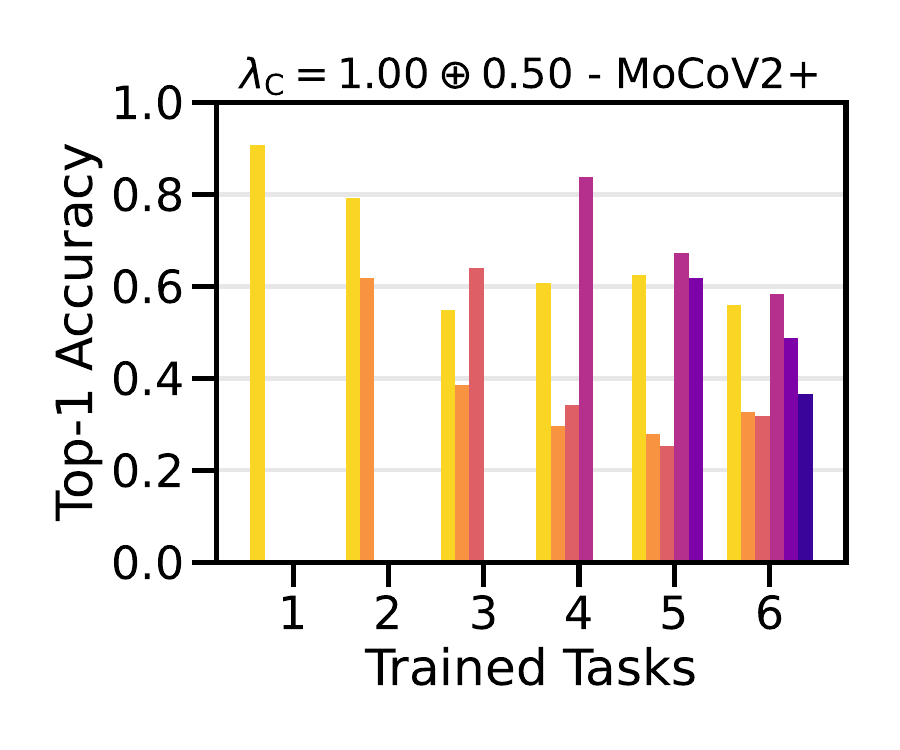}
    \includegraphics[width=0.24 \linewidth ]{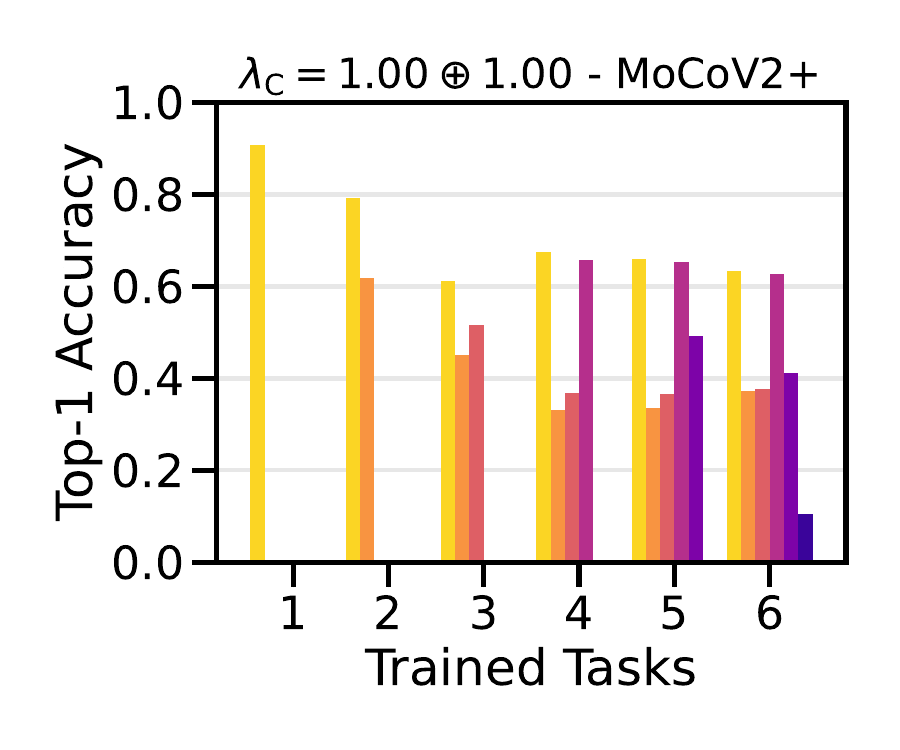} \\
    \vspace{-0.05in}
    \includegraphics[width=0.65 \linewidth ]{figures_new/Part_1/F4-WISDM2019-6Tasks-v2_legend.pdf}
    \vspace{-0.1in}
    
    \caption{Detailed breakdown of performance over tasks with progressive importance coefficients. Fine-grained accuracy is shown for every additional task with different importance coefficients.}
    \vspace{-0.2in}
    \label{fig:kaizen_performance_per_task_progressive_lamb}
\end{figure}

We conducted 8 additional experiments with different progressive importance coefficients: $\lambda_{\mathrm{C}}$ = $0.50 \oplus 0.00$, $0.50 \oplus 0.25$, $0.50 \oplus 0.50$, $0.50 \oplus 1.00$, $1.00 \oplus 0.00$, $1.00 \oplus 0.25$, $1.00 \oplus 0.50$, $1.00 \oplus 1.00$, and the results are shown in Fig.~
 \ref{fig:kaizen_general_performance_progressive_lamb} and \ref{fig:kaizen_performance_per_task_progressive_lamb} (additional visualisation can be found in Appx.~\ref{subsection:additional_results}).
From the detailed breakdown (see Fig.~\ref{fig:kaizen_performance_per_task_progressive_lamb}), we can verify our understanding: the performance on trained tasks remains almost unchanged at the last few tasks when the progressive factor is high (see the plots of $0.50 \oplus 1.00$ and $1.00 \oplus 1.00$). The use of a progressive factor allows the model to shift the focus from new task learning to knowledge retention over time. The results shown in Fig. \ref{fig:kaizen_general_performance_progressive_lamb} further support our hypothesis on proportional importance: with $\lambda_{\mathrm{C}}=0.50 \oplus 0.50$, which corresponds to scaling the importance of knowledge retention proportional to the number of tasks learned, the model can achieve the highest average continual performance and final performance. A higher value of $\lambda_{\mathrm{C}}=1.00 \oplus 1.00$ can still maintain high performance, but an increased coefficient at later steps hurts the performance of the final model. Other settings display a spectrum of performance with varying priorities given to knowledge distillation and current task learning. This set of results allows us to better explore the trade-off between knowledge retention and new task learning, a key aspect of continual learning.

%% file: sections/5-conclusion.tex
\section{Conclusion}
In this work, we adapted two state-of-the-art CSSL frameworks, CaSSLe and Kaizen, from visual representation learning to human activity recognition, one of the fundamental tasks in human-centric computing. Our evaluation indicates that a unified training scheme handling both representation learning and classification learning, as proposed in Kaizen, can perform better under realistic data assumptions, with the advantage of being deployable at any point during the process, which is particularly vital for HAR and other human-centric applications. Additional experiments indicate that the use of a progressive importance coefficient which adaptively adjusts the importance of knowledge retention and classification learning can allow us to explore the trade-off between different learning objectives, reaching higher levels of performance compared to a fixed loss function. This work demonstrated the potential of utilising self-supervised learning for developing human activity recognition models that can adapt to changes in user behaviours.

%% file: sections/6-appendix.tex
\begin{appendix}
    \input{sections/2-related}
\section{Transformation Functions}
\label{appx:trans}

\noindent\textbf{Random 3D rotation} applies a random rotation in the 3D space by picking a random axis in 3D and a rotational angle from uniform distributions. This is to simulate common pose changes in wearable devices.

\noindent\textbf{Random scaling} alters the size of samples within a window by multiplying them with a randomly chosen scalar. We apply this transformation since a model that can handle these scaled signals creates better representations because it learns to be unaffected by changes in amplitude and offset.

\noindent\textbf{Time warping} locally stretches or warps the time series data, smoothly distorting the time intervals between sensor readings.
\section{Evaluation setup}
\label{appx:setup}
{
\subsection{Model Architecture}
In this work, we adopted a lightweight HAR model, TPN \cite{multi_self_har}, to replace the vision-based model.

\subsection{Evaluation Metrics}
For the evaluation metrics, we are adopting the same framework as introduced in Kaizen \cite{tang2023practical}: \textbf{Final Accuracy (FA)}, \textbf{Continual Accuracy (CA)}, \textbf{Forgetting (F)}, and \textbf{Forward Transfer (FT)} as metrics. This set of values can better reflect the performance of continual learning methods in different use cases.

\subsection{Self-supervised Learning Frameworks}
In this work we selected a contrastive-based method MoCoV2+~\cite{chen2020improved, he2020momentum}, and an asymmetric-model-based method
BYOL~\cite{grill2020bootstrap} as the self-supervised learning method for continual learning. These two methods have been shown to be well-performing in continual self-supervised learning settings~\cite{fini2022self, tang2023practical}, and our goal is to investigate whether different CSSL methods demonstrate different performance characteristics when using different SSL methods as the knowledge retention mechanism.

\subsection{Dataset}
We performed our evaluation using the WISDM2019 (WISDM Smartphone and Smartwatch Activity and Biometrics Dataset) \cite{weiss2019wisdm}, which is an activity recognition dataset collected by the WISDM (Wireless Sensor Data Mining) Lab in the Department of Computer and Information Science of Fordham Unversity. The dataset contains raw accelerometer and gyroscope data from a smartwatch (LG G Watch) and a smartphone (Google Nexus 5/5x or Samsung Galaxy S5) worn by 51 subjects, who performed 18 different activities for 3 minutes each. The smartphone is placed inside the participant's pocket, while the smartwatch is worn at the dominant hand. The data was collected at a sampling rate of 20Hz for the following activities: Walking, Jogging, Stairs, Sitting, Standing, Typing, Brushing Teeth, Eating Soup, Eating Chips, Eating Pasta, Drinking from Cup, Eating Sandwich, Kicking (Soccer Ball), Playing Catch with Tennis Ball, Dribbling (Basketball), Writing, Clapping, and Folding Clothes. The accelerometer data from the smartwatch was used in this study, and we selected this dataset for evaluation because it has a relatively high number of activities in HAR, which makes it suitable for continual learning evaluation.

The raw sensor data underwent minimal pre-processing. First, z-normalization was applied using the training data's mean and standard deviation for each sensor channel. Next, the data was segmented into \(384 \times 3\) sliding windows, representing 384 timestamps and 3 triaxial accelerometer channels. Consecutive windows do not overlap. 20-25\% of user data was held out unseen as the test set to evaluate model generalisability. 

The 18 classes of activities are randomly and evenly split into 6 tasks of 3 classes each with no overlap as follows: task 1 - \{Folding Clothes, Stairs, Walking\}, 
task 2 - \{Sitting, Drinking from Cup, Eating Chips\}, 
task 3 - \{Standing, Eating Sandwich, Clapping\}, 
task 4 - \{Brushing Teeth, Jogging, Eating Pasta\},
task 5 - \{Eating Soup, writing, Typing\}, and
task 6 - \{Playing Catch with Tennis Ball, Kicking (Soccer Ball), Dribbling (Basketball)\}. This follows the conventional setup of class-incremental learning, and how different (potentially qualitative) splitting of the classes affects model performance is left as future work.

\subsection{Baselines}
\label{subsection:baselines}
We followed the evaluation setup as proposed in \cite{tang2023practical}, in which we compare the performance of Kaizen, CaSSLe, as well as a \emph{No distill} baseline that fine-tunes the full model on new tasks with no explicit catastrophic forgetting mitigation. As CaSSLe and \emph{No distill} baselines are self-supervised learning methods without incorporating a classifier, the classifiers are trained separately after the feature extractor is trained, with data replay enabled.
}
\end{appendix}

\section{Additional Visualisations}
\label{subsection:additional_results}
\begin{figure}
    \centering
    \includegraphics[width=0.8 \linewidth ]{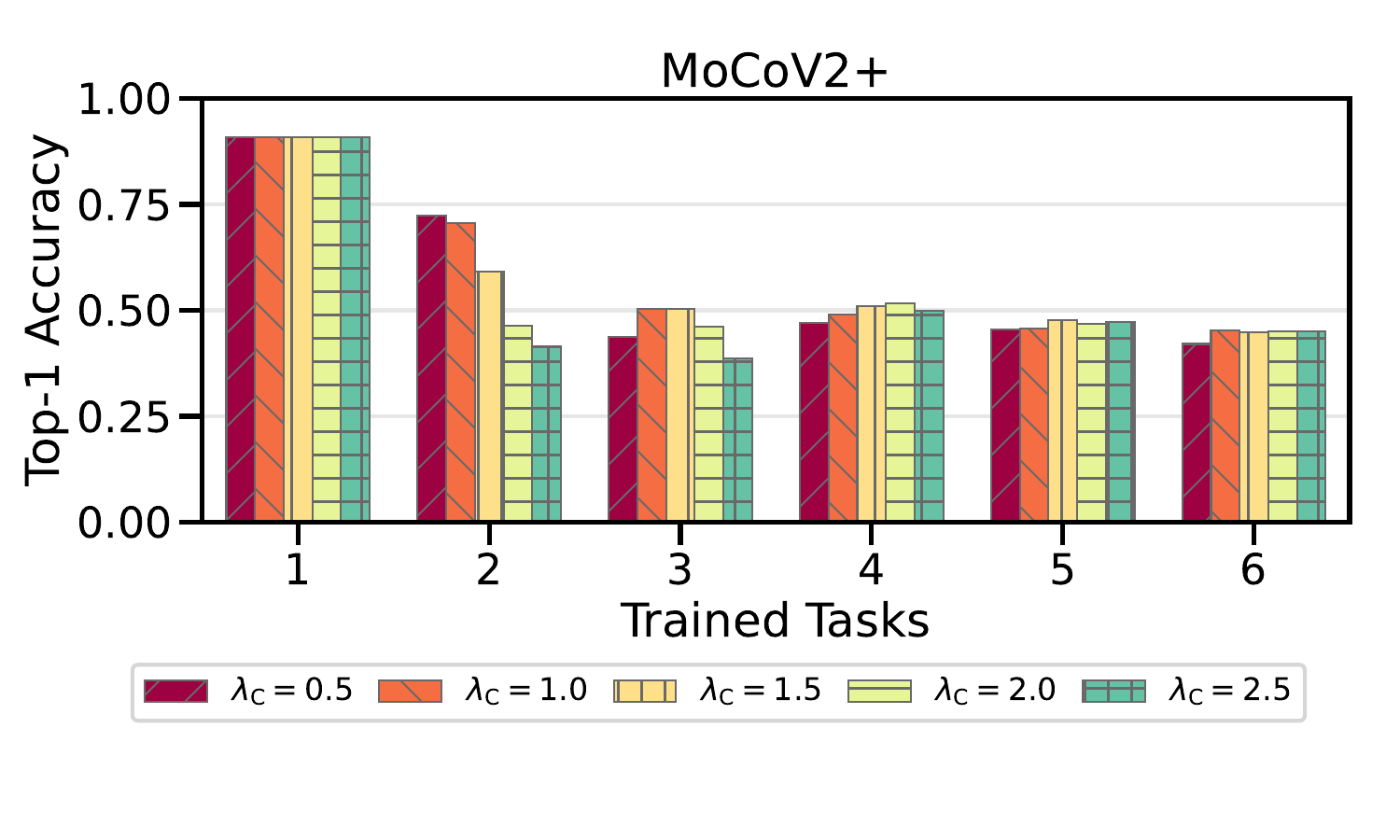}
    \caption{Performance after training on different tasks with varying constant importance coefficients.}
\label{fig:kaizen_performance_across_time_constant_lamb}
\end{figure}

\begin{figure}
    \centering
    \includegraphics[width=0.99 \linewidth ]{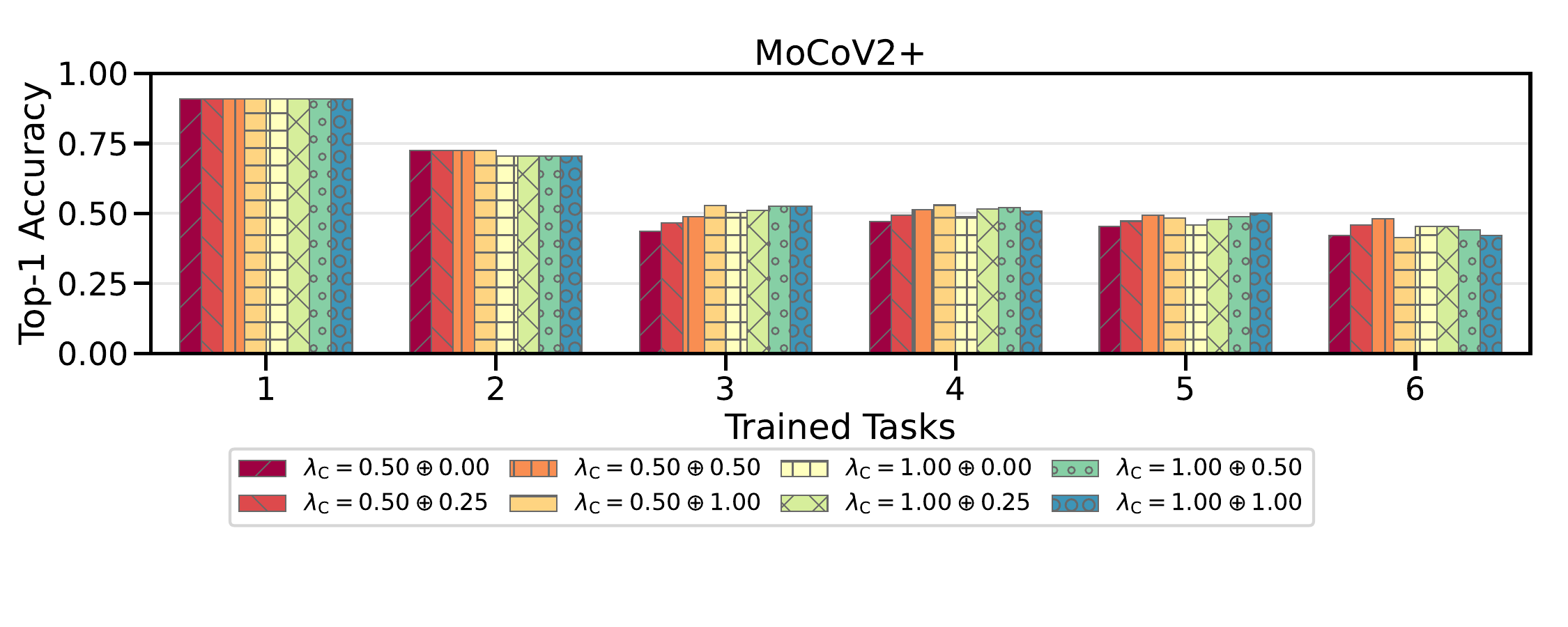}
    \caption{Performance after training on different tasks with varying progressive importance coefficients.}
\label{fig:kaizen_performance_across_time_progressive_lamb}
\end{figure}

Fig.~\ref{fig:kaizen_performance_across_time_constant_lamb} and \ref{fig:kaizen_performance_across_time_progressive_lamb} illustrate the aggregated performance of models trained with Kaizen after each task with different importance coefficients.

%% file: sections/2-related.tex
\section{Related Work}
\label{related}
Continual learning has been an active area of research in which many researchers have proposed techniques for mitigating catastrophic forgetting. The main idea behind these methods is to balance stability (retaining knowledge) and plasticity (learning new concepts) for deep learning models. Techniques broadly fall into three categories~\cite{de2021continual}: regularisation-based~\cite{kirkpatrick2017overcoming, li2017learning, shin2017continual, wu2019large}, replay-based~\cite{rebuffi2017icarl, chaudhry2018efficient, ostapenko2019learning, buzzega2020dark}, and parameter-isolation methods~\cite{rusu2016progressive, serra2018overcoming}. State-of-the-art performance often combines these approaches~\cite{mittal2021essentials}.

\subsection{Continual Self-supervised Learning}
However, many existing techniques rely heavily on labelled data, which is often unavailable.  Continual self-supervised learning (CSSL) approaches leverage self-supervised learning to enable continual learning under limited supervision. Initial efforts focused narrowly on self-supervised pre-training combined with supervised continual learning~\cite{gallardo2021self, caccia2022special} or extending contrastive learning frameworks~\cite{cha2021co2l, madaan2021rethinking}. However, these techniques are often designed to work with specific self-supervised learning frameworks. The latest continual self-supervised learning frameworks proposed in existing literature~\cite{de2021continual, fini2022self} have begun investigating more overarching and flexible frameworks, where the model learns continually from a stream of unlabelled data. The work by \cite{tang2023practical} proposed a unified semi-supervised framework with continual fine-tuning, introducing mechanisms for knowledge distillation and new task learning in both representation and classification. These demonstrate progress towards practical solutions under realistic supervision assumptions, and we focus on these recent proposals which repurpose SSL methods for continual learning in this work.

\subsection{Human Activity Recognition}
Wearable-based human activity recognition is an important component in human-centric computing because of its ability to extract real-time information and context clues about user behaviours, which enables other computing applications \cite{choi2016understanding, jaimes2015corredor}. However, the utility of HAR systems has been limited by the scarcity of high-quality labelled data and the computational capabilities of wearable devices.

Since the adoption of deep learning models for HAR, researchers have looked at tackling catastrophic forgetting in HAR specifically \cite{jha2021continual, leite2022resource, schiemer2023online}, and they have demonstrated moderate success. Simultaneously, self-supervised learning has recently gained popularity in the wearable-based HAR community, and many approaches from the general machine learning community, as well as customised methods, have been proposed to leverage unlabelled data to overcome the limitations of labelled data \cite{multi_self_har, tang2020exploring, haresamudram2020masked, tang2021selfhar, haresamudram2021contrastive}. However, we have yet to see efforts in leveraing self-supervised learning techniques for HAR in continual learning. This is an important step towards practical and generalisable user models. CSSL paves the way for realistic solutions to continually learn from combinations of labelled and unlabelled data.